\UseRawInputEncoding
\documentclass[letterpaper, 10 pt, journal, twoside]{IEEEtran} 

\title{Safety-Critical and Distributed Nonlinear Predictive Controllers for Teams of Quadrupedal Robots}

\usepackage{amsmath,amssymb,amsfonts}
\usepackage{cite}
\usepackage{graphicx} 
\usepackage{epsfig} 
\usepackage{times} 
\usepackage{xcolor}
\usepackage{balance}
\usepackage{multicol}
\usepackage{array}
\usepackage{multirow}
\usepackage{booktabs}
\usepackage[colorlinks=true, allcolors=blue]{hyperref}

\newcommand{\Real}{\mathbb{R}}
\newcommand{\col}{\textrm{col}}

\newcommand{\identity}{\mathbb I}

\newcounter{definitionCounter}
\newcounter{theoremCounter}
\newcounter{lemmaCounter}

\newcommand{\definition}[1]{
  \refstepcounter{definitionCounter} 
  \vspace{0.5em}
  \noindent\textbf{\textit{Definition} \textit{\thedefinitionCounter}} \textit{(#1)}: 
}

\author{Basit Muhammad Imran$^{1}$,  Jeeseop Kim$^{2}$, Taizoon Chunawala$^{1}$, Alexander Leonessa$^{1}$, and Kaveh Akbari Hamed$^{1}$

\thanks{The work of B. Imran was supported by the National Science Foundation (NSF) under Grant 1924617. The work of K. Akbari Hamed is supported by the NSF under Grants 2024772 and 2423725.}
\thanks{$^{1}$B. Imran, T. Chunawala, A. Leonessa, and K. Akbari Hamed are with the Department of Mechanical Engineering, Virginia Tech, Blacksburg, VA 24061, USA, {\tt\small \{basit, taizoonc, aleoness, kavehakbarihamed\}@vt.edu}}
\thanks{$^{2}$J. Kim is with California Institute of Technology, Pasadena, CA 91125, USA, {\tt\small jeeseop@caltech.edu}}
}

\begin{document}
\maketitle

\begin{abstract}
This paper presents a novel hierarchical, safety-critical control framework that integrates distributed nonlinear model predictive controllers (DNMPCs) with control barrier functions (CBFs) to enable cooperative locomotion of multi-agent quadrupedal robots in complex environments.  While NMPC-based methods are widely adopted for enforcing safety constraints and navigating multi-robot systems (MRSs) through intricate environments, ensuring the safety of MRSs requires a formal definition grounded in the concept of invariant sets. CBFs, typically implemented via quadratic programs (QPs) at the planning layer, provide formal safety guarantees. However, their zero-control horizon limits their effectiveness for extended trajectory planning in inherently unstable, underactuated, and nonlinear legged robot models. Furthermore, the integration of CBFs into real-time NMPC for sophisticated MRSs, such as quadrupedal robot teams, remains underexplored. This paper develops computationally efficient, distributed NMPC algorithms that incorporate CBF-based collision safety guarantees within a consensus protocol, enabling longer planning horizons for safe cooperative locomotion under disturbances and rough terrain conditions. The optimal trajectories generated by the DNMPCs are tracked using full-order, nonlinear whole-body controllers at the low level. The proposed approach is validated through extensive numerical simulations with up to four Unitree A1 robots and hardware experiments involving two A1 robots subjected to external pushes, rough terrain, and uncertain obstacle information. Comparative analysis demonstrates that the proposed CBF-based DNMPCs achieve a 27.89\% higher success rate than conventional NMPCs without CBF constraints.
\end{abstract}

\begin{IEEEkeywords}
Legged robots, motion control, multi-contact whole-body motion planning and control
\end{IEEEkeywords}


\vspace{-1em}
\section{Introduction}

Multi-robot systems (MRSs), especially those consisting of legged robots, play a crucial role in mission-critical tasks due to their exceptional capability to traverse rough terrains. Applications include collaborative firefighting, unmanned rescue, disaster response, and exploration. Recent advancements in legged robotics have significantly enhanced these robots' agility and robustness, enabling them to navigate complex terrains more effectively, see, e.g., \cite{Kim_Wensing_Convex_MPC_01,Hutter_anymal_cbf_inWBC,MIT_humanoid_cbf_inWBC}. While multi-robot systems have been extensively studied (see, e.g., \cite{CBF_MRS,MRS_CBF_Cavorsi}), MRSs involving legged robots pose unique challenges due to their high dimensionality, unilateral constraints, underactuation, and hybrid nature. These factors distinguish them from other MRSs and may limit the applicability of existing control approaches, particularly for quadrupedal robots.


\begin{figure}
    \centering
    \includegraphics[width=0.85\linewidth]{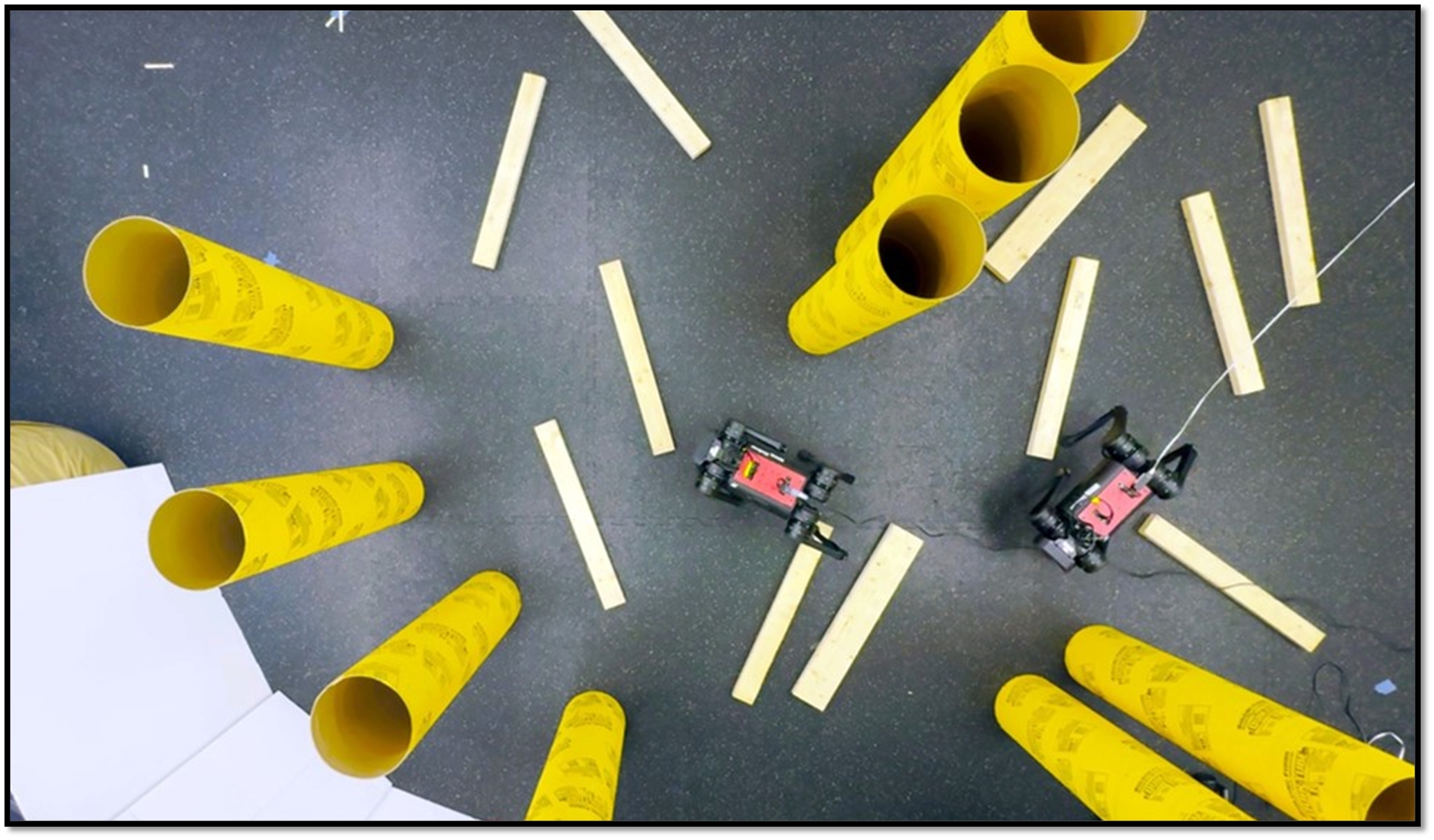}
    \vspace{-1.1em}
    \caption{Top-view snapshot of an experiment demonstrating CBF-based DNMPC algorithms, where two Unitree A1 robots navigate a challenging environment with uncertainties in obstacle positions and rough terrain.} \vspace{-1.5em}
    \label{Fig_Sanpshot}
\end{figure}


Reduced-order models (i.e., templates) \cite{Full_Koditschek_Template} provide low-dimensional representations of complex, nonlinear locomotion systems and can be effectively integrated with model predictive control (MPC) for real-time trajectory planning in legged robots. Common reduced-order models include the linear inverted pendulum (LIP) model \cite{kajita19991LIP} and its variants, such as angular momentum LIP (ALIP) \cite{ALIP}, spring-loaded inverted pendulum (SLIP) \cite{SLIP}, vertical SLIP (vSLIP) \cite{vLIP_Sreenath}, hybrid LIP (H-LIP) \cite{HLIP_Ames},  centroidal dynamics \cite{orin2013centroidal}, and the single rigid body (SRB) model \cite{Kim_Wensing_Convex_MPC_01,Wensing_VBL_HJB,Abhishek_Hae-Won_TRO}. For single-agent robots, MPC algorithms based on linearized templates reduce to convex quadratic programs (QPs), which can be efficiently solved in real-time \cite{pandala2022robust,Leila_Hamed_RAL}. However, these QP-based MPCs struggle to handle the nonlinear and nonconvex constraints required for collision avoidance in MRSs, necessitating the use of nonlinear MPC (NMPC) \cite{Basit_ASME}. NMPC has also been employed for gait planning in single-agent quadrupedal robots, e.g., \cite{NMPC_Park_02,NMPC_CBF_Ames_Hutter,NMPC_CBF_Sreenath,Hutter_anymal_cbf_inWBC, MIT_humanoid_cbf_inWBC}. The previous work \cite{Basit_ASME} explored distributed NMPC (DNMPC)-based planning for multi-agent quadrupedal robots, addressing nonconvex robot-to-robot and robot-to-obstacle collision avoidance using LIP models but without considering robot orientation. While NMPC can handle such constraints, ensuring MRS safety requires a formal framework grounded in invariant sets and control barrier functions (CBFs) \cite{Ames_CBF,CBF_MRS,Nonsmooth_CBF,MRS_CBF_Cavorsi}. 


\begin{figure*}[t]
    \centering
    \includegraphics[width=0.9\linewidth]{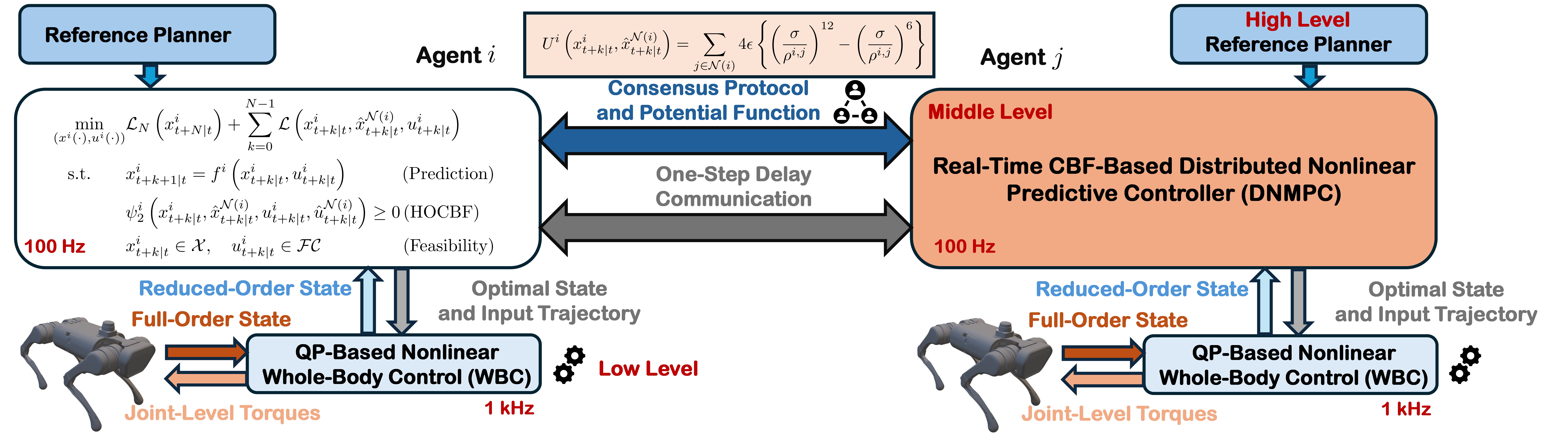}
    \vspace{-1.0em}
    \caption{Overview of the proposed hierarchical control based on CBF-based distributed NMPCs for safe locomotion of multi-agent quadrupedal robots.}\vspace{-1.5em}
    \label{Fig_Overview}
\end{figure*}


CBF-based controllers for MRSs are formulated as QP-based optimization problems that modify nominal controllers while enforcing barrier constraints \cite{CBF_MRS,MRS_CBF_Cavorsi}. These QPs can be viewed as optimal control problems for input-affine systems with a zero-control horizon. To enable the safe and cooperative locomotion of two quadrupedal robots,  \cite{JKim_CBF_CooperativeLoco} proposed a hierarchical framework where a high-level QP-based safety controller with CBFs generates reduced-order reference trajectories, which are then tracked by a low-level QP-based MPC for gait generation. Notably, CBFs and MPC operate in separate layers rather than within a unified framework. Recent studies \cite{Koushil_CBFMPC, CBFNMPC_Koushil} have explored integrating CBFs within MPC and NMPC frameworks, combining the benefits of a nonzero prediction horizon with simple integrator dynamics. Additionally, \cite{NMPC_CBF_Sreenath} employed CBFs with duality-based obstacle avoidance constraints in an NMPC framework running at 30 Hz for a single-agent quadrupedal robot. Moreover, \cite{NMPC_CBF_Ames_Hutter} implemented a low-frequency safe kinematic MPC with first-order CBFs for single-agent quadrupedal robots. However, safety guarantees within a distributed NMPC framework for complex multi-agent legged robots remain unexplored.

The \textit{overarching goal} of this work is to develop computationally efficient, unified distributed NMPC algorithms that integrate CBF-based collision safety guarantees with a consensus protocol, thereby enabling extended planning horizons for the safe locomotion of multi-agent quadrupedal robots in the presence of disturbances and challenging terrain conditions.


\vspace{-0.8em}
\subsection{Related Work}

MPC implementations for MRSs can generally be categorized into centralized, decentralized, and distributed approaches. Centralized control offers optimal performance but is often impractical for large-scale or communication-constrained systems due to its high computational and communication requirements \cite{Jeeseop_TRO}. In contrast, decentralized MPC assigns independent controllers to each agent without communication, reducing complexity but sacrificing effectiveness in tightly coupled systems \cite{Siljak_Decentralized_Book}. Distributed MPC (DMPC) offers a compromise by allowing limited inter-agent communication to balance individual performance with collective behavior \cite{2009_Scattolini_DMPC}. DMPC can be further classified into cooperative and non-cooperative frameworks. Cooperative DMPC optimizes a shared global objective to enhance system-wide coordination \cite{venkat2005stability, rawlings2008coordinating}, whereas non-cooperative DMPC focuses on local objectives. Various DMPC algorithms have been developed to tackle specific scenarios, such as systems with coupled constraints but decoupled dynamics \cite{richards2007robust} or those requiring iterative information sharing \cite{camponogara2002distributed}.
Non-cooperative DMPC strategies have become increasingly popular for collision avoidance in MRSs, characterized by local cost functions with decoupled dynamics and coupled constraints. These strategies have been applied to autonomous vehicles \cite{Abdelaal2019}, differential drive robots \cite{non-convex-PB-DMPC}, and linear systems \cite{FARINA20121088}.

While the literature on DMPC is extensive, extending these approaches to distributed nonlinear MPC (DNMPC) for multi-agent quadrupedal robots presents significant \textit{challenges} due to the inherent instability of locomotion models, underactuation, and unilateral constraints. Previous work \cite{Basit_ASME} developed a DNMPC approach for collision-safe navigation in quadrupedal MRSs by directly incorporating Euclidean distance-based constraints into the NMPC framework of LIP models. While effective, this approach faces limitations in computational tractability, with the real-time NMPC loop constrained to just 5.5 Hz. In contrast, \cite{JKim_CBF_CooperativeLoco} proposed collision safety by imposing CBFs on the high-level path planner. However, applying CBFs at this level ensures safety only for the reference path given to the NMPC, not for the real-time NMPC itself. Recent work \cite{Koushil_CBFMPC} has attempted to integrate CBFs with NMPC, but this was limited to simple integrator dynamics. 


\vspace{-1em}
\subsection{Contributions}

This paper aims to develop a unified, computationally efficient, and fast distributed control framework that integrates DNMPC and CBFs to enable the safe locomotion of multi-agent quadrupedal robots. The \textit{contributions} of the paper are as follows. The paper introduces a hierarchical control framework comprising three layers (see Fig. \ref{Fig_Overview}). At the middle layer, we propose a real-time distributed NMPC algorithm operating at 100 Hz to address trajectory optimization of nonlinear SRB models of multi-agent quadrupedal robots. The DNMPC incorporates discrete-time CBF constraints to ensure robot-to-robot and robot-to-obstacle safety while leveraging a Lennard-Jones potential function in the cost formulation of NMPCs for consensus protocol. Reference trajectories for the DNMPC are generated offline by a high-level planner using potential fields combined with the $A^{\star}$ algorithm. The optimal trajectories computed by the middle-layer DNMPC are subsequently provided to low-level nonlinear whole-body controllers (WBCs) operating at 1 kHz, which 
impose the full-order model of each agent to track the optimal trajectories. The WBCs are developed based on QP and virtual constraints \cite{Jessy_Book}.  

We validate the proposed formulation through experiments on two Unitree A1 robots and numerical simulations involving a team of four A1 robots navigating uncertain environments with obstacle location uncertainty, rough terrain, and external disturbances, as shown in Fig. \ref{Fig_Sanpshot}. The results demonstrate a significant improvement in collision avoidance success rates, increasing from 66\% in previous approaches \cite{Basit_ASME} to 94\% with our method. Additionally, we achieve a substantial improvement in computational efficiency, raising the nonlinear predictive control loop update rate from 5.5 Hz in the previous work \cite{Basit_ASME} to 100 Hz with the proposed CBF-based DNMPC framework.  To further evaluate the efficacy of the unified CBF-based DNMPC approach compared to DNMPC algorithms employing separate CBF conditions at the low level, we perform numerical simulations on randomly generated terrains with randomly positioned obstacles. The results show a notable 79.39\% improvement in performance with the unified CBF-based DNMPC approach.

The current work differs from the previous work \cite{Jeeseop_TRO,Randy_ICRA_MultiAgent}, which focused on the cooperative locomotion of quadrupedal robots with holonomic constraints for payload transport using convex QP-based predictive controllers. Unlike those works, this study employs NMPCs and explicitly addresses safety and collision avoidance, which were not considered in  \cite{Jeeseop_TRO,Randy_ICRA_MultiAgent}. Furthermore, the work is different from \cite{Multi_Agent_Quadrupeds_Sreenath}, which focuses on manipulating cable-towed loads using a cascaded planning scheme combining parallelized centralized and decentralized planners at 0.5 and 20 Hz, based on duality \cite{Borrelli_Opt_Coll_Avoi} rather than CBFs. In contrast, our approach utilizes entirely 100 Hz CBF-based distributed NMPCs, rather than centralized MPCs.


\vspace{-0.5em}
\section{Problem Formulation}
\label{se:two}

This section presents the problem formulation by addressing local safety sets for individual agents within the MRS. We consider a network of $n_{A}$ quadrupedal agents, indexed by the set  $\mathcal{A}:= \{1,\cdots,n_A\}$. Variables associated with each agent $i\in\mathcal{A}$ are denoted using the superscript $i$. The discrete-time nonlinear dynamics of each agent $i\in\mathcal{A}$ are described by the following state equation  
\begin{equation} \label{eqn:nonlinear-system-dt}
    x^i(t+1)=f^{i}\left(x^i(t), u^i(t)\right),
\end{equation}
where $t\in\mathbb{Z}_{\geq 0}:= \{0,1,\cdots\}$ represents the discrete time, $x^{i}\in\mathcal{X}$ and $u^{i}\in\mathcal{U}$ denote the local state variables and local control inputs, respectively, and $\mathcal{X}\subset\Real^{n}$ and $\mathcal{U}\subset\Real^{m}$ represent the feasible state and the admissible control sets, for some positive integers $n$ and $m$. The function $f^{i}:\mathcal{X}\times\mathcal{U}\rightarrow\mathcal{X}$ is continuous. The global state and control input vectors are defined as $x:=\col\{x^{i}\,|\,i\in\mathcal{A}\}\in\Real^{nn_{A}}$ and $u:=\col\{u^{i}\,|\,i\in\mathcal{A}\}\in\Real^{mn_{A}}$, where ``$\col$'' denotes the column operator. 

\textbf{Collision Safety in the MRS:} Each quadrupedal agent navigates a complex environment with static obstacles while ensuring inter-robot and obstacle avoidance. Static obstacles are represented by the Cartesian coordinates of their center points, denoted as $o^\ell\in\Real^{2}$ for $\ell \in \mathcal{O}:=\{1,\cdots,n_{O}\}$, where $n_{O}$ represents the number of obstacles. To address safety, we assume that the center of mass (COM) coordinates of each agent in the xy-plane are given by $g(x^{i})$, where $g:\Real^{n}\rightarrow2$ is a continuous mapping. We next define two global safe sets as super-level sets of the Euclidean distance:  $\mathcal{SA}$, capturing safe distances between all agents, and $\mathcal{SO}$, ensuring safe distances between agents and obstacles. More specifically, we define
\begin{alignat}{4}
    &\!\!\mathcal{SA}&&:=\{x\in\Real^{nn_{A}}|\|g(x^{i})-g(x^{j})\|\geq d_{\textrm{th}}, \forall i\neq j \in \mathcal{A}\}\nonumber\\
    &\!\!\mathcal{SO}&&:=\{x\in\Real^{nn_{A}}|\|g(x^{i})-o^{\ell}\|\geq d_{\textrm{th}}, \forall i\in\mathcal{A}, \forall \ell\in\mathcal{O}\},
\end{alignat}
where $\|\cdot\|$ denotes the Euclidean distance, and $d_{\textrm{th}}$ is a threshold distance value. The overall \textit{global safe set} $\mathcal{S}$ is then given by the intersection $\mathcal{S}:=\mathcal{SA} \cap \mathcal{SO}$.

Ensuring system-wide safety can be formulated as maintaining the forward invariance of $\mathcal{S}$ \cite{Ames_CBF}. However, designing a centralized NMPC law to enforce this invariance is computationally demanding. To mitigate this, we propose local NMPC laws that ensure the safety of individual agents by defining \textit{local safety sets}. Specifically, for each agent $i\in\mathcal{A}$, two local safe sets can be defined accordingly:  $\mathcal{SA}^{i}$ for capturing safe distances between agent $i$ and any other agent $j\neq i \in \mathcal{A}$, and $\mathcal{SO}^{i}$ for ensuring safe distances between agent $i$ and obstacles $o^{\ell}$ for all $\ell\in\mathcal{O}$, that is,
\begin{align}
    \mathcal{SA}^{i} &:= \bigcap_{j\neq i\in\mathcal{A}} \left\{ x^i \in \mathcal{X}\, \middle| \, \|g(x^{i}) -  g(x^{j})\| \geq d_{\textrm{th}} \right\} \label{eqn:dynamic_safe_set} \\
     \mathcal{SO}^{i} &:= \bigcap_{\ell\in\mathcal{O}} \left\{ x^i \in \mathcal{X} \, \middle| \, \|g(x^{i}) - o^\ell\| \geq d_{\textrm{th}} \right\}. \label{eqn:static_safe_set}
\end{align}   
The combined local safe set $\mathcal{S}^i$ is then obtained by the intersection of these two sets as $\mathcal{S}^{i} := \mathcal{SA}^{i} \cap \mathcal{SO}^{i}$.

The local safe set can be reformulated as 
\begin{align} \label{eqn:safe-set}
    \mathcal{S}^{i} = \left\{x^{i} \in \mathcal{X}\, \middle| \, h^{i}(x^{i}, x^{\mathcal{N}(i)}, o) \geq 0\right\},
\end{align}
where $\mathcal{N}(i):=\{j\in\mathcal{A}\,| \,j\neq i\}$ denotes the set of neighboring agents for agent $i$, $x^{\mathcal{N}(i)}:=\col\{x^{j}\,|\,j\in\mathcal{N}(i)\}\in\Real^{n(n_{A}-1)}$ represents the states of all neighboring agents, and $o:=\col\{o^{\ell}\,|\,\ell\in\mathcal{O}\}\in\Real^{2n_{O}}$ contains the coordinates of all obstacles. In addition, $h^{i}$ is a continuous local function that can be expressed as follows:
\begin{equation}\label{eqn:loal_func}
    h^{i}(x^{i}, x^{\mathcal{N}(i)}, o):=\begin{bmatrix}
    \col\left\{h^{i,j}\,|\,j\in\mathcal{N}(i)\right\}\\
    \col\left\{h^{i,\ell}\,|\,\ell\in\mathcal{O}\right\}
    \end{bmatrix},
\end{equation}
where $h^{i,j}:=\|g(x^{i}) - g(x^{j})\|-d_{\textrm{th}}$ for all $j\in\mathcal{N}(i)$ and $h^{i,\ell}:=\|g(x^{i})-o^{\ell}\|-d_{\textrm{th}}$ for all $\ell\in\mathcal{O}$. Since $h^{i}$ depends on the local state $x^{i}$ as well as the states of other agents $x^{\mathcal{N}(i)}$, the local safety set $\mathcal{S}^{i}$ is parameterized by the states of all neighboring agents at every time step $t$, i.e., $\mathcal{S}^{i}=\mathcal{S}^{i}(x^{\mathcal{N}(i)})$. 

\textbf{Problem Statement:} We aim to design computationally efficient, distributed NMPC laws that compute the optimal local control law $u^{i}\in\mathcal{U}$ for each agent with a consensus protocol while (1) ensuring the invariance of its local safe set $\mathcal{S}^{i}$, and (2) maintaining a consistent ``sticking distance'' between the agents. To account for the dependence of local safety sets on other agents' states, we will adopt a communication protocol in which agents share their optimal state and input trajectories---computed by their local NMPCs---with a one-step delay. This enables each local NMPC to optimize its own state and control trajectories independently without solving for those of other agents.

Before establishing the proposed DNMPC framework, we present definitions that introduce the concept of higher-order, discrete-time CBFs for each agent. For notational simplicity, we will use $x_{t}^{i}$ and $u_{t}^{i}$ to denote $x^{i}(t)$ and $u^{i}(t)$, respectively.

\definition{Discrete-Time CBF \cite{DT-HOCBF}}
The local function $h^{i}$ is said to be a CBF for \eqref{eqn:nonlinear-system-dt} if there exists class $\mathcal{K}$ function $\alpha$ \cite{Khalil_Book} satisfying $\alpha(s) < s$ for all $s > 0$ such that
\begin{equation}
    \Delta h^{i}\left(x^{i}(t), u^{i}(t)\right) \geq - \alpha\left(h^{i}(x^{i}(t))\right), \quad \forall x^{i}(t) \in \mathcal{X},
\end{equation}
where $\Delta h^{i}(x^{i}(t), u^{i}(t)) := h^{i}(x^{i}(t+1)) - h^{i}(x^{i}(t))$ and $s$ is the argument of $\alpha$.

For systems with a higher relative degree $r > 1$, the control input does not appear in the first-order difference of the CBF $h^{i}$. In this case, a series of functions can be defined based on the original $h^{i}(x_t^i)$ with the relative degree $r$ as follows:
\begin{align} \label{eqn:hocbf-function-candidates}
    \psi^{i}_{0}(x_t^i) &:= h^{i}(x_t^i) \nonumber \\
    \psi^{i}_{1}(x_t^i) &:= \Delta \psi^{i}_{0}\left(x_t^i, u_t^i\right) + \alpha_1\left(\psi^{i}_{0}(x_t^i)\right) \nonumber \\
    &\vdots \nonumber \\
    \psi^{i}_{r}(x_t^i,u_t^i) &:= \Delta \psi^{i}_{r-1}\left(x_t^i, u_t^i\right) + \alpha_r\left(\psi^{i}_{r-1}(x_t^i)\right),
\end{align}
where $\Delta\psi^{i}_{a}(x_{t}^{i},u_{t}^{i}):=\psi^{i}_{a}(x^{i}_{t+1})-\psi^{i}_{a}(x^{i}_{t})$ for $0\leq a \leq r-1$ and $\alpha_{a}$ are class $\mathcal{K}$ functions satisfying $\alpha_{a}(s)<s$ for all $a=1,\cdots,r$ and every $s>0$. This series of functions yields a corresponding series of sets, as follows: 
\begin{align}\label{eqn:hocbf-candidate-sets}
    \mathcal{S}^{i}_{0} &:= \left\{ x_t^i \in \mathcal{X}\, \middle|\, \psi^{i}_{0}(x_t^i) \geq 0 \right\} \nonumber \\
    \mathcal{S}^{i}_{1}&:= \left\{ x_t^i \in \mathcal{X} \,\middle|\, \psi^{i}_{1}(x_t^i) \geq 0 \right\} \nonumber \\
    &\vdots \nonumber \\
    \mathcal{S}^{i}_{r-1} &:= \left\{ x_t^i \in \mathcal{X} \,\middle|\, \psi^{i}_{r-1}(x_t^i) \geq 0 \right\}. 
\end{align}

\definition{Higher-Order Discrete-Time CBF \cite{DT-HOCBF}} \label{def_HOCBF}
The local function $h^{i}$ is a higher-order, discrete-time CBF (HOCBF) of relative degree $r$ if there exist functions $\psi^{i}_{a}$ for $a \in \{0, 1, \hdots, r\}$ defined by \eqref{eqn:hocbf-function-candidates} and corresponding sets $\mathcal{S}_{a}^{i}$ for $a \in \{0, 1, \hdots, r-1\}$ defined by \eqref{eqn:hocbf-candidate-sets} such that 
\begin{equation}\label{eqn:hocbf-condition}
    \psi_{r}^{i}(x_{t}^{i},u_{t}^{i}) \geq 0
\end{equation}
for all $x_{t}^{i} \in \bigcap_{a=0}^{r-1} \,\mathcal{S}^{i}_{a}$ and for some control input $u^{i}_{t}\in\mathcal{U}$.

The following theorem formally establishes a sufficient condition on the local control law  $u^i$ to ensure that the intersection set $\bigcap_{a=0}^{r-1}\,\mathcal{S}_{a}^{i}$ remains invariant for local dynamics. 

\begin{theorem} \textit{(HOCBF Condition \cite{DT-HOCBF}):}\label{Thm_HOCBF}
If $h^{i}$ is a continuous HOCBF of relative degree $r$ defined on $\bigcap_{a=0}^{r-1}\,\mathcal{S}_{a}^{i}$, any control input $u^{i}(t)\in\mathcal{U}$ satisfying the HOCBF condition \eqref{eqn:hocbf-condition} will render $\bigcap_{a=0}^{r-1}\,\mathcal{S}_{a}^{i}$ forward invariant for agent $i\in\mathcal{A}$. 
\end{theorem}


\vspace{-0.5em}
\section{Distributed NMPCs based on HOCBFs}
\label{se:three}

This section proposes the middle-level distributed NMPC framework based on HOCBFs for real-time trajectory planning to enable the safe locomotion of a team of multi-agent quadrupedal robots in environments with obstacles. Additionally, the section addresses a consensus protocol incorporated into the cost function of local NMPCs. 


\textbf{Template Model:} The SRB dynamics of each agent are used as the template model, with the state variables defined as the COM position and orientation, along with their time derivatives, that is, $x^{i}:=\col(p^{i},\dot{p}^{i},\Theta^{i},\omega^{i})\in\Real^{n}$ with $n=12$, where $p^{i}\in\Real^{3}$ represents the Cartesian coordinates of the COM of agent $i\in\mathcal{A}$, $\Theta^{i}\in\Real^{3}$ denotes the Euler angles (roll, pitch, and yaw) of the body, and $\omega^i\in\Real^{3}$ represents the angular velocities in the body frame. The control inputs $u^{i}(t)$ are further taken as the ground reaction forces (GRFs) that generate the net force and torque around the COM, denoted by $(f^{i,\textrm{net}},\tau^{i,\textrm{net}})$ and expressed in the world frame. The equations of motion can then be described by 
\begin{equation}\label{eqn:SRB-dynamics-ct}
    \Sigma^i: \begin{cases}
    \ddot{p}^i      = \frac{{f}^{i,\text{net}}}{m_{\textrm{tot}}} - g_0 \\
    \dot{\Theta}^i  = A(\Theta^{i})\,\omega^i \\
    \dot{\omega}^i  = I^{-1} \left( R^i {}^{\top} \tau^{i,\text{net}} - \mathbb{S}(\omega^i) \, I\, \omega^i \right),
\end{cases}
\end{equation}
where $m_{\textrm{tot}}$ is the total mass of the agent, $I\in\Real^{3\times3}$ represents the moment of the inertia in the body frame, $g_{0}\in\Real^{3}$ denotes the gravitational constant,  $R^{i}\in\textrm{SO}(3)$ represents the orientation matrix of the body frame relative to the world frame, and $\mathbb{S}(\cdot):\Real^3 \rightarrow \mathfrak{so}(3)$ is the skew-symmetric operator. The net force and torque around the COM are then calculated as a linear combination of the individual GRFs acting at each contacting leg. Specifically, a time-varying matrix 
$E(t)$ maps the vector of individual GRFs to the net wrench as follows:
\begin{equation}
    \begin{bmatrix}
        f^{i,\textrm{net}}\\
        \tau^{i,\textrm{net}}
    \end{bmatrix}=E(t)\,u^{i}(t).
\end{equation}
The nonlinear SRB dynamics in \eqref{eqn:SRB-dynamics-ct} can be discretized and expressed in the form of \eqref{eqn:nonlinear-system-dt} by the Euler method. Furthermore, the model remains valid if the GRFs (i.e., inputs $u^{i}(t)$) belong to the friction cone, denoted by $\mathcal{U}=\mathcal{FC}$. 

\textbf{Collision Safety for Quadrupedal Robots}: From Definition \ref{def_HOCBF}, local functions $h^{i}$ in \eqref{eqn:loal_func} serve as continuous HOCBF candidates with relative degree $r=2$ if there exist a series of functions $\psi^{i}_{1}, \psi^{i}_{2}$, and corresponding sets $ \mathcal{S}_{0}^{i}, \mathcal{S}^{i}_{1}$  such that 
\begin{equation} \label{eqn:hocbf-nmpc-condition}
    \psi^{i}_{2}\left(x_t^i, x_t^{\mathcal{N}(i)}, u_t^i, u_t^{\mathcal{N}(i)}\right) \geq 0, \quad \forall x_t^i \in 
\mathcal{S}^{i}_{0} \cap \mathcal{S}^{i}_{1}.
\end{equation}
We remark that $\psi^{i}_{2}$ generally depends on the local state and input variables $(x_t^i,u_t^i)$, as well as the variables of all other agents, i.e., $(x_t^{\mathcal{N}(i)},u_{t}^{\mathcal{N}(i)})$. Furthermore, from Theorem \ref{Thm_HOCBF}, any feasible local control input (i.e., GRFs) $u^{i}(t)\in\mathcal{U}$ satisfying the HOCBF condition \eqref{eqn:hocbf-nmpc-condition} will render $\mathcal{S}^{i}_{0} \cap \mathcal{S}^{i}_{1}$ invariant. With the conditions for multi-agent safety now established, we proceed to formulate a distributed NMPC framework embedding the safety condition \eqref{eqn:hocbf-nmpc-condition} into its constraints.

\textbf{HOCBF-based DNMPCs:} To address the problem in Section \ref{se:two}, we propose the following DNMPC formulation for each agent $i\in\mathcal{A}$:
\begin{alignat}{2} \label{eqn:hocbf-nmpc-formulation}
  &\min_{(x^{i}(\cdot),u^{i}(\cdot))} && \mathcal{L}_N\left(x_{t+N|t}^{i}\right) + \sum_{k=0}^{N-1} \mathcal{L} \left(x_{t+k|t}^{i}, \hat{x}_{t+k|t}^{\mathcal{N}(i)}, u_{t+k|t}^{i}\right) \nonumber\\
  &\quad \text{s.t.} \, && x^{i}_{t+k+1|t}=f^{i}\left(x^{i}_{t+k|t},u^{i}_{t+k|t}\right) \quad \quad \quad \textrm{(Prediction)}\nonumber\\
  & && \psi_2^{i}\left(x_{t+k|t}^{i}, \hat{x}_{t+k|t}^{\mathcal{N}(i)}, u_{t+k|t}^{i}, \hat{u}^{\mathcal{N}(i)}_{t+k|t}\right) \geq 0 \, \textrm{(HOCBF)} \nonumber\\
  & && x^i_{t+k|t} \in \mathcal{X}, \quad u_{t+k|t}^{i} \in \mathcal{FC} \qquad \qquad \,\,\,\, \textrm{(Feasibility)},
\end{alignat} 
where $N$ is the control horizon, $x^{i}_{t+k|t}$ and $u^{i}_{t+k|t}$ represent the predicted local states and local inputs at time $t+k$, computed at time $t$ using the prediction model with the initial condition of $x^{i}_{t|t}=x^{i}(t)$. The terms $\hat{x}^{\mathcal{N}(i)}_{t+k|t}$ and $\hat{u}^{\mathcal{N}(i)}_{t+k|t}$
denote the \textit{estimation} of predicted variables of all neighboring agents for agent $i$, obtained using the \textit{one-step delay communication protocol (OSDCP)}, which will be described later. The inequality constraints in the local NMPC \eqref{eqn:hocbf-nmpc-formulation} represent the HOCBF condition \eqref{eqn:hocbf-nmpc-condition} together with the feasibility conditions for the states and inputs (i.e., GRFs). The local NMPC optimizes the state and input trajectories of agent $i$, denoted by $(x^{i}(\cdot),u^{i}(\cdot))$. 

\textbf{Consensus Protocol:} The cost function of the local NMPC consists of the terminal and stage costs. The terminal cost is taken as $\mathcal{L}_{N}(x^{i}_{t+N|t}):=\|x^{i}_{t+N|t}-x^{i,\textrm{ref}}_{t+N|t}\|_{P}^{2}$, where $x^{i,\textrm{ref}}(\cdot)$ denotes the local reference trajectory generated by the high-level planner (see Section \ref{se:four}), and $P=P^{\top}$ is a positive definite matrix.  The stage cost is defined as
\begin{alignat}{4}
&\mathcal{L} \left(x_{t+k|t}^{i}, \hat{x}_{t+k|t}^{\mathcal{N}(i)}, u_{t+k|t}^{i}\right)&&:=
\|x^{i}_{t+k|t}-x^{i,\textrm{ref}}_{t+k|t}\|_{Q}^{2}+\|u_{t+k|t}\|_{R}^{2} \nonumber\\
& && \,+ w\, U^{i}\left(x_{t+k|t}^{i}, \hat{x}_{t+k|t}^{\mathcal{N}(i)}\right),\label{eqn:local_cost}
\end{alignat}
where $Q=Q^{\top}$ and $R=R^{\top}$ are positive definite matrices, $w>0$ is a weighting factor, and $U^{i}$ is an artificial potential function for consensus. The potential function $U^{i}$, inspired by Lennard-Jones potential model \cite{LJ1924}, is given by
\begin{equation}\label{eqn:LJ_potential}
    U^{i}\left(x_{t+k|t}^{i}, \hat{x}_{t+k|t}^{\mathcal{N}(i)}\right):= \sum_{j\in\mathcal{N}(i)} 4\epsilon\left\{\left(\frac{\sigma}{\rho^{i,j}}\right)^{12} - \left(\frac{\sigma}{\rho^{i,j}}\right)^{6}\right\},
\end{equation}
where $\rho^{i,j}:=\|p^{i}_{\textrm{x,y}}-\hat{p}^{j}_{\textrm{x,y}}\|$ denotes the estimated distance between the COMs of two agents $i$ and $j\neq{i}$, $\hat{p}^{j}_{\textrm{x,y}}$ represents the estimate of the COM position of agent $j$, and $\epsilon$ and $\sigma$ are tuning parameters. Incorporating the potential function $U^{i}$ serves as a \textit{consensus protocol}, enabling agents to maintain proximity through van der Waals-like interaction potential forces. Specifically, the parameter $\sigma$ defines the zero-potential distance and is tuned to achieve the equilibrium distance at which the interaction force is zero. By including this potential function in the cost of the local NMPC, agents are encouraged to maintain a consistent ``sticking distance'', effectively emulating soft-holonomic constraints on the distances between their COMs. In the experiments section, we demonstrate that this consensus protocol successfully emulates soft-flocking behavior among the agents while leveraging the safety guarantees provided by the HOCBF-DNMPC framework. By soft-flocking behavior, we refer to non-strict constraints, meaning that agents can occasionally deviate from these soft requirements to prioritize meeting hard requirements, such as ensuring collision safety.

\textbf{OSDCP Protocol: \label{OSDCP}} The one-step delay communication protocol (OSDCP) is proposed to address the inherent challenge of solution unavailability from other agents in \eqref{eqn:hocbf-nmpc-formulation}. Specifically, due to the distributed nature of the formulation, the optimal solution $(x^{j,\star},u^{j,\star})$, of agent $j$, computed at time $t$,  is not immediately accessible by agent $i$. Consequently, the condition in \eqref{eqn:hocbf-nmpc-condition} cannot be directly applied. To facilitate implementation, the OSDCP leverages the optimal solutions computed by other agents at the preceding time step ($t-1$). With a sufficiently small time step $T_{s}$, agent $i$ approximates the predicted states and inputs of all other agents $j \in\mathcal{N}(i)$ as
$\hat{x}_{k+t|t}^{j} \approx x^{j, \star}_{k+t|t-1}$ and $\hat{u}_{k+t|t}^{j} \approx u^{j, \star}_{k+t|t-1}$, where $(x^{j, \star}_{k+t|t-1},u^{j,\star}_{k+t|t-1})$
denotes the optimal predicted variables of agent $j$ from the previous time step $(t-1)$.


\vspace{-0.5em}
\section{High-level and Low-level Layers}
\label{se:four}

This section presents a concise overview of the high- and low-level layers of the control scheme, adapted from the previous works \cite{Basit_ASME,Randy_Paper_LCSS}, with modifications.

\textbf{High-Level Path Planner:} The proposed hierarchical control framework incorporates an offline high-level path planner (see Fig. \ref{Fig_Overview}) that generates reference trajectories for the middle-layer DNMPC algorithms. To achieve this, we employ a hybrid path planning approach that combines the $A^{\star}$ algorithm \cite{MARTINS2022e01068} with the potential fields (PFs) method \cite{spong2005robot}. The previous work \cite{Basit_ASME} relied solely on a PF-based path planner. However, it is well-known that PFs are prone to local minima issues. Similar to the work in \cite{hybrid_A-star_APF}, we adopt the $A^\star$ algorithm to generate 
globally optimal way-points, which are then bridged by locally smooth and dynamics-informed reference paths using PFs. The reference trajectories are then passed to the middle-layer DNMPC algorithm. These distributed predictive control algorithms track the reference trajectories using the SRB model of each agent while making real-time adjustments to address disturbances, rough terrains, and environmental uncertainties, ensuring compliance with the HOCBF constraints. 
 

\textbf{Low-Level Nonlinear WBC:} At the low level, we use nonlinear WBCs to impose the full-order dynamical model of each quadrupedal robot for tracking the optimal reduced-order state and GRF trajectories prescribed by the middle-layer DNMPCs. The WBC, adopted from the previous work \cite{Randy_Paper_LCSS}, is formulated as a real-time QP running at 1 kHz. It incorporates the tracking problem within the framework of virtual constraints \cite{Jessy_Book} for the floating-base full-order model while solving for dynamically feasible joint-level torques.


\begin{figure*}[t]
    \centering
    \includegraphics[width=1.00\linewidth]{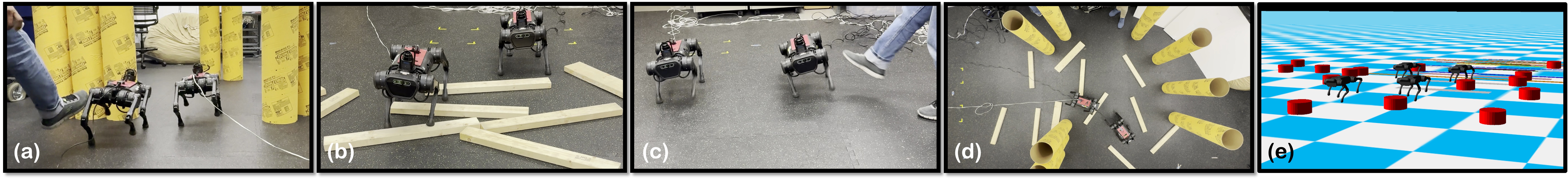}
    \vspace{-2em}
    \caption{Snapshots of experiments demonstrating the deployment of CBF-based DNMPCs in a multi-agent A1 setup under various conditions: (a) uncertain obstacles, (b) intentional intersecting reference trajectories on rough terrain, (c) straight-line reference trajectories with one agent physically pushed onto another, and (d) cooperative locomotion on rough terrain with uncertain obstacles. (e) Snapshot of simulations with four agents navigating rough terrain.}
    \vspace{-1.5em}
    \label{Fig_Snaphosts}
\end{figure*}


\vspace{-0.8em}
\section{Experiments}
\label{se:five}

This section presents the results of our numerical simulations and hardware experiments, thoroughly assessing the proposed distributed control algorithm's effectiveness.


\vspace{-1em}
\subsection{Setup and Distributed Controller Synthesis}

This work utilizes A1 quadrupedal robots as a test platform for the numerical and experimental validation of the proposed framework. The A1 robots are compact quadrupeds with a mass of 12.45 (kg), height of 0.28 (m), and 18 degrees of freedom (DOFs), 12 of which are actuated through three active joints per leg: hip pitch, hip roll, and knee pitch. The robots are equipped with an NVIDIA Jetson TX2, an RP-LiDAR S1 operating at 15 Hz for point-cloud generation and obstacle detection, and an onboard computer with an Ethernet switch. The distributed algorithms are executed on multi-threads of an offboard desktop PC with an AMD Ryzen 7950X3D CPU and 64 GB of DDR5 RAM. For real-time obstacle detection, we utilize the obstacle detector ROS package \cite{obstacle_detector}. Numerical simulations are conducted in the RaiSim physics engine \cite{RAISIM}, employing four A1 robots in the simulation environment, while two are deployed for hardware evaluations.

The parameters of the CBF-based DNMPCs are set as $Q=\textrm{block diag}\{Q_{p},Q_{\dot{p}},Q_{\Theta},Q_{\omega}\}$ with $Q_{p}=\textrm{diag}\{1e5, 1e5, 8e6\}$, $Q_{\dot{p}}=\textrm{diag}\{5e5, 5e5, 8e6\}$, $Q_{\Theta}=\textrm{diag}\{1e4, 1e4, 1e4\}$, and $Q_{\omega}=\textrm{diag}\{1e4, 1e4, 1e4\}$, $P=100\,Q$ and $R=\identity$. The control horizon for the local NMPCs is chosen as $N=10$ with the sampling time of $T_{s}=10$ (ms). Additional hyperparameters include a collision avoidance threshold distance of $d_{\textrm{th}}=0.6$ (m), class $\mathcal{K}$ functions of $\alpha_{1}(s)= 0.1\,s$ and $\alpha_{2}(s)=0.05\,s$ for HOCBF, the tuning parameters of the Lennard-Jones potential in \eqref{eqn:LJ_potential} as $\epsilon = 50$, $\sigma = 0.85$ (m), and a weighting factor of $w=1e9$ in \eqref{eqn:local_cost}. We choose trotting gaits of quadrupedal robots with a step time of 180 (ms) and the local nonlinear programs (NLPs) in \eqref{eqn:hocbf-nmpc-formulation} include 240 decision variables. The real-time distributed NMPCs are solved using CasADi \cite{CasADI} framework along with IPOPT solver with 8 iterations, utilizing the previous solution as the initial guess. We evaluate the framework in a complex environment with up to $n_{O}=20$ obstacles, as illustrated in Fig. \ref{Fig_Snaphosts}. As the number of agents and obstacles increases in experiments, the continuous function $h^{i}$ in \eqref{eqn:safe-set} can be reformulated as
\begin{equation}
    h^{i}:=\min\{\min_{j\in\mathcal{N}(i)}h^{i,j},\min_{\ell\in\mathcal{O}}h^{i,\ell}\},
\end{equation}
which significantly reduces the number of inequality constraints in the NLPs derived from the HOCBF. This reformulation results in an NMPC computation time of approximately $4.9$ (ms) for two and four agents. The low-level QP-based nonlinear WBC is solved for torque control at 1 kHz \cite{Randy_Paper_LCSS}. 


\vspace{-1em}
\subsection{Hardware Experiments and Numerical Simulations}

To validate the efficacy of the proposed approach, we conducted experiments using two A1 robots in a multi-agent configuration under various challenging scenarios, as shown in Fig. \ref{Fig_Snaphosts}. Videos are available online \cite{YouTube_CBF_DNMPC}. 

\textbf{Modifications of trajectories by DNMPC:} In the first experiment, the robots were commanded to follow intersecting trajectories to assess collision avoidance on rough terrain. This setup specifically evaluated the performance of middle-level DNMPCs when handling colliding reference paths. Instead of relying on a high-level planner, both robots were provided with $x$ and $y$ velocity references, leading to an intentional collision scenario. Additionally, wooden blocks were used to simulate rough terrain with a maximum height of 5 (cm) (17.85\% of the robots' standing height). Under these conditions, the CBF-based DNMPC algorithm successfully adjusted the reference commands to prevent collisions, as shown in Fig. \ref{Fig_Snaphosts}(b).

\textbf{Safety with disturbances (external push):} The second experiment aimed to evaluate the effectiveness of the distributed control framework by applying an external push to one agent and recording its response. Both agents were assigned straight-line forward reference trajectories without lateral deviations. During execution, one agent was physically perturbed to simulate an unexpected collision event. The consensus protocol, combined with the CBFs in the middle-level DNMPC, enabled the unperturbed agent to react promptly by moving away from the collision zone, maintaining a constant separation distance between the agents, as shown in Fig. \ref{Fig_Snaphosts}(c).


\begin{figure}
    \centering
    \includegraphics[width=1.0\linewidth]{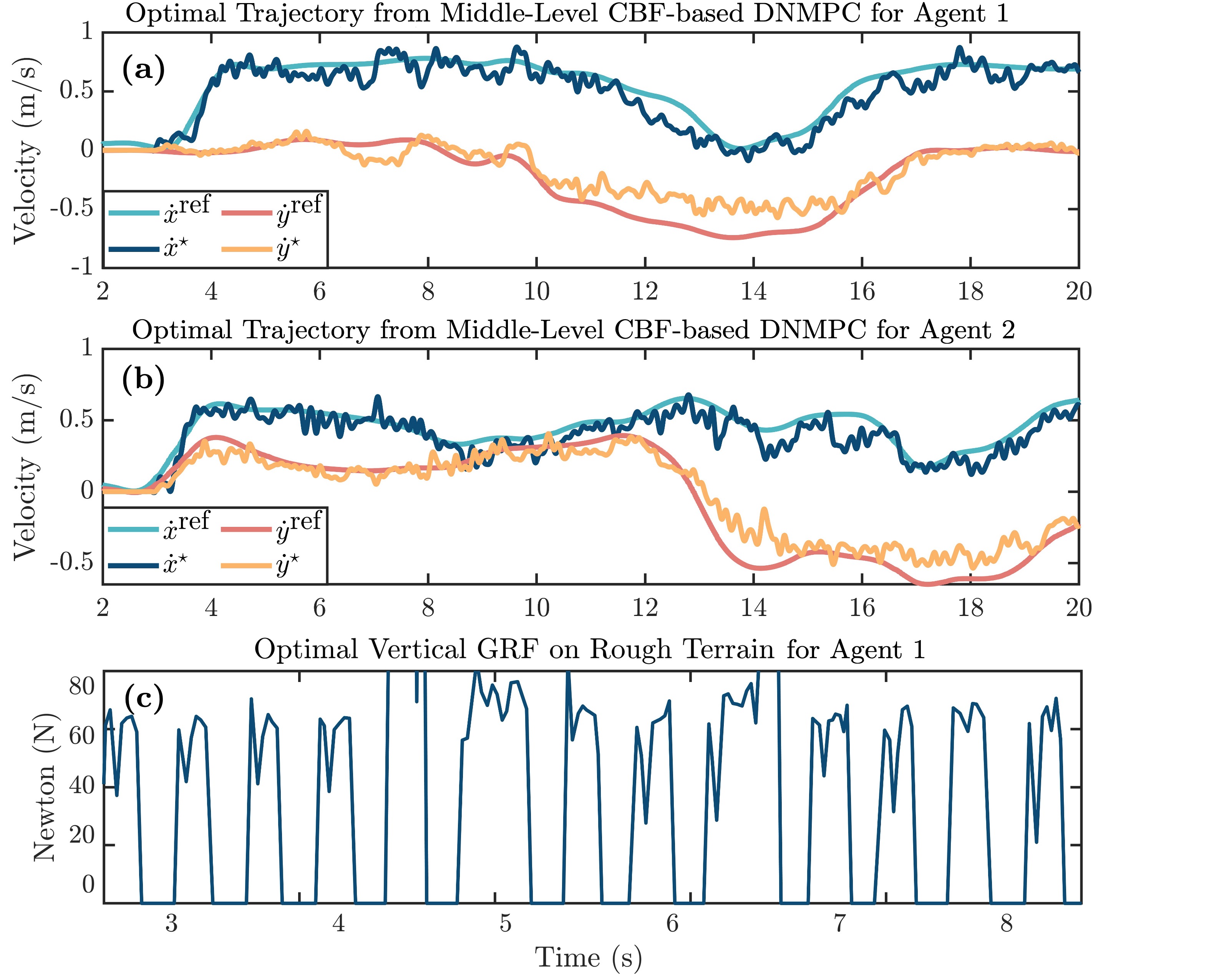}
    \vspace{-2.2em}
    \caption{Plot of the reference COM velocity trajectories  ($\dot{x}^\textrm{ref}, \dot{y}^\textrm{ref}$) generated by the high-level planner along with the optimal velocity trajectories ($\dot{x}^{\star}, \dot{y}^{\star}$) adjusted by the middle-level DNMPCs for agents 1 and 2 in (a) and (b) during experiments of Fig. \ref{Fig_Snaphosts}(d). (c) Plot of the vertical component of the GRF profile prescribed by the DNMPC for agent 1. } \vspace{-1.8em}
    \label{Fig_Opt_Trajectories}
\end{figure}


\textbf{Robustness against uncertainty and rough terrain:} The third scenario was designed to assess the robustness of CBF-based DNMPCs in environments with uncertain obstacle information (see Fig. \ref{Fig_Snaphosts}(d)). Initially, the high-level planner operated under the assumption of known obstacle locations. However, during execution, slight perturbations were deliberately introduced to the obstacle positions, creating a mismatch between the assumed and actual environment. The high-level planner first generated the desired trajectories, after which the middle-level DNMPCs were activated to account for the perturbed obstacles. Additionally, wooden blocks were introduced to evaluate the system’s robustness under both rough terrain and obstacle uncertainty. Figure \ref{Fig_Opt_Trajectories} illustrates the velocity trajectories prescribed by the high-level planner and the corresponding optimal trajectories generated by DNMPCs for agents 1 and 2. The plot also shows the optimal GRFs computed by the local NMPC for the front left leg of agent 1.

\begin{figure}
    \centering
    \includegraphics[width=1.0\linewidth]{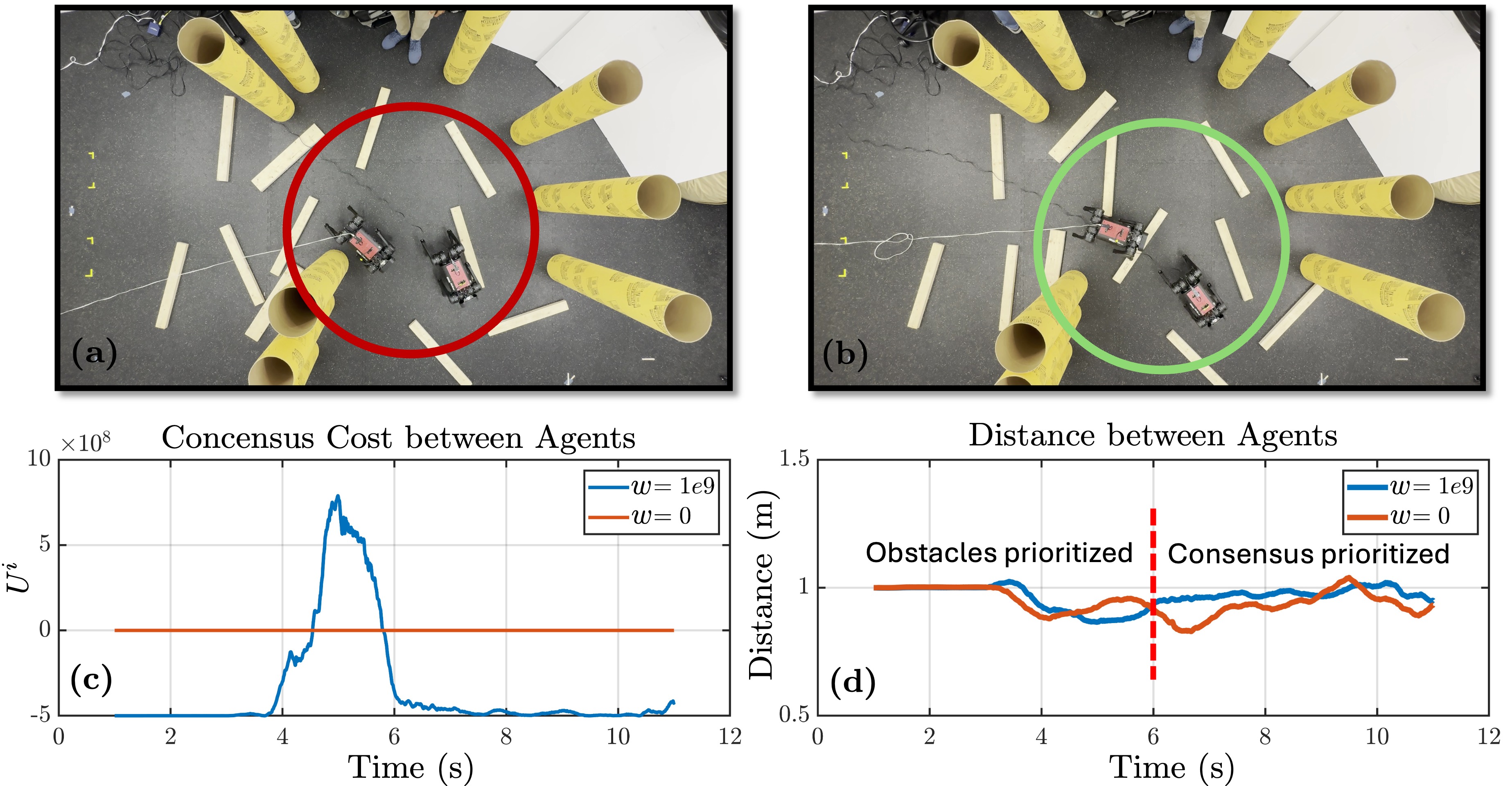}
    \vspace{-2.2em}
    \caption{Snapshots of experiments (a) with consensus protocol and (b) without consensus protocol. (c) Plots for the consensus cost $w U^i$ for $w = 0$ (no consensus) and $w=10^9$ (with consensus). (d) Distance between the two A1 agents for $w = 0$ and $w=10^9$, where it can be seen that after the consensus has been achieved, agents maintain a sticking distance when obstacle avoidance is not an active constraint in the DNMPC.} \vspace{-1.1em}
    \label{Fig_flocking_behavior}
\end{figure}

\textbf{Consensus protocol:} The efficacy of the consensus protocol in \eqref{eqn:LJ_potential} and \eqref{eqn:local_cost} was empirically evaluated through a controlled experiment involving two A1 robots navigating nine uncertain obstacles over wooden blocks. Two experimental configurations were tested: one with the consensus protocol deactivated ($w=0$) and another with a high-gain activation ($w=1e9$). Figure \ref{Fig_flocking_behavior} demonstrates that activating the protocol improved soft-flocking behavior and enhanced inter-agent cohesion. Figure \ref{Fig_flocking_behavior}(c) presents the actual cost of the consensus term for both cases, offering insight into the degree of coordination under different protocol settings. Figure \ref{Fig_flocking_behavior}(d) depicts the inter-agent distance over time, revealing deviations from the nominal 1 (m) separation when CBF constraints are active ($t<6$ (s)), indicating collision avoidance maneuvers. Notably, when the CBF constraints are inactive ($t>6$ (s)), the enabled consensus protocol maintains a more stable inter-agent distance.

\textbf{Simulations with four agents:} We further evaluated the framework's performance in RaiSim using four A1 robots navigating wooden blocks and 20 randomly distributed obstacles, as shown in Fig. \ref{Fig_Snaphosts}(e). The agents successfully and safely traversed the complex terrain. The simulation video is available online \cite{YouTube_CBF_DNMPC}. Notably, no infeasibility issues were observed with the DNMPCs during the simulations or experiments. The following section presents a quantitative analysis based on numerical simulations.


\vspace{-1em}
\subsection{Comparison and Quantitative Analysis}

\begin{figure}
    \centering
    \includegraphics[width=0.7\linewidth]{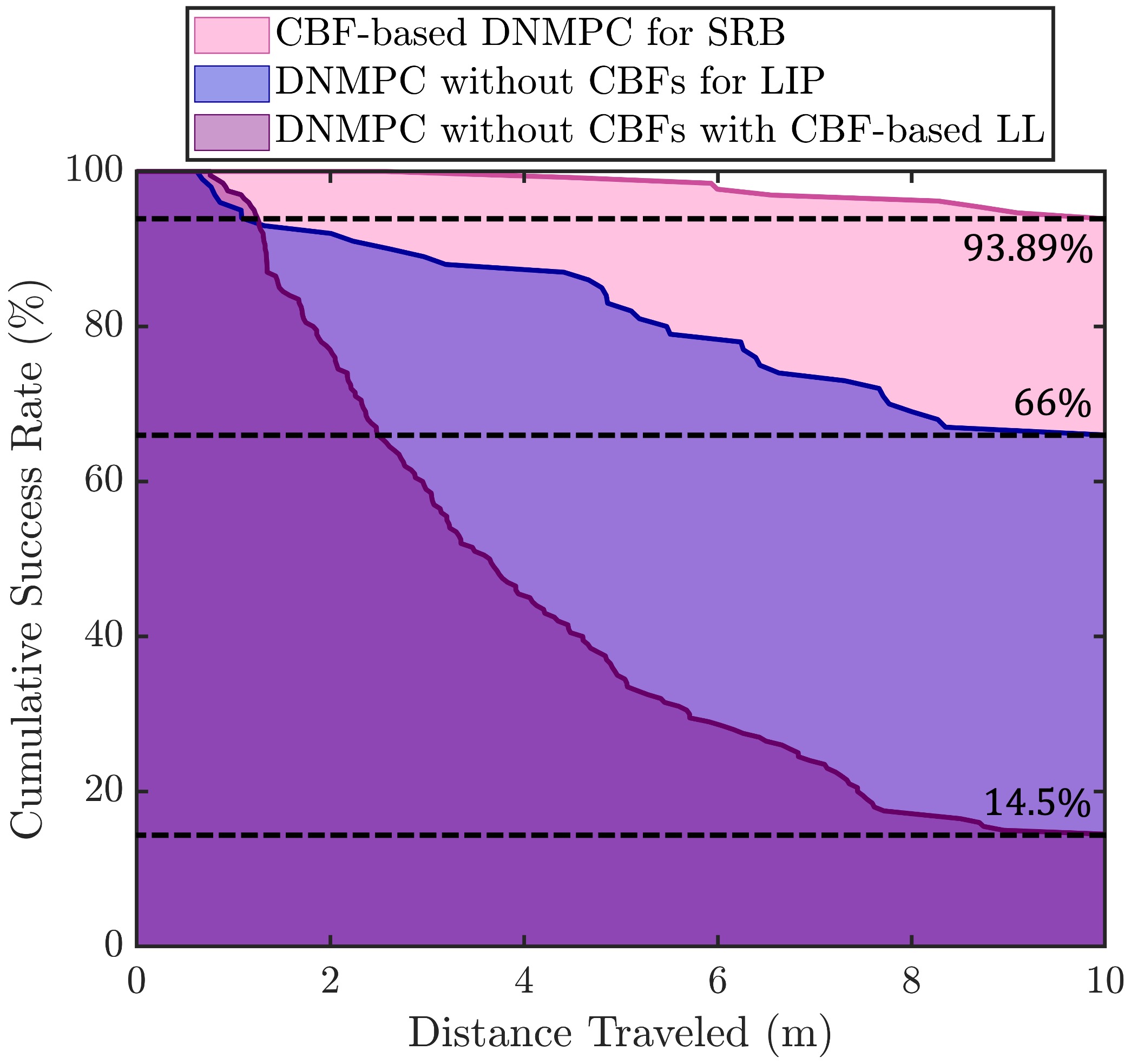}
    \vspace{-1.1em}
    \caption{Success rate comparison of the proposed CBF-based DNMPC for SRB against two alternative approaches: (i) DNMPC without CBFs but incorporating Euclidean distance constraints for LIP and (ii) DNMPC without CBFs but utilizing CBF-based WBCs.} 
    \vspace{-1.8em}
    \label{Fig_comparison_plot}
\end{figure}

We conduct extensive numerical simulations to evaluate the efficacy and robustness of the proposed CBF-based DNMPC algorithms compared to other distributed predictive control techniques. The evaluation focuses on the cooperative and safe locomotion of two A1 quadrupedal agents trotting across 200 randomly generated terrains with 20 randomly positioned obstacles. For comparison, we employ three different middle-level distributed predictive controllers while maintaining the same high- and low-level controllers. The first layered controller implements the proposed CBF-based DNMPC algorithms applied to SRB models. The second layered controller uses DNMPC algorithms without CBF conditions, adopted from \cite{Basit_ASME}, and applied to LIP dynamics. In this case, safety is enforced through Euclidean distance constraints for collision avoidance without incorporating CBFs into the DNMPCs. Finally, the third layered controller employs DNMPC algorithms with SRB dynamics but without CBF conditions, where CBF constraints are enforced at the low-level WBCs instead of the middle-level NMPCs, similar to the typical WBC approach used for single-agent robots, see, e.g., \cite{nguyen20163d, Hutter_anymal_cbf_inWBC, MIT_humanoid_cbf_inWBC}. Figure \ref{Fig_comparison_plot} illustrates the success rate of different local NMPC approaches for agent one as a function of traveled distance. Here, success is defined as both agents reaching 10 (m) without robot-to-robot or robot-to-obstacle collisions and maintaining gait stability. The overall success rates of the proposed CBF-based DNMPC for SRB, DNMPC without CBFs but with Euclidean distance constraints for LIP, and DNMPC without CBFs but with CBF-based WBCs are 93.8\%, 66\%, and 14.5\%, respectively, demonstrating a significant performance improvement with respect to the previous work \cite{Basit_ASME}. 


\vspace{-1em}
\section{Conclusions}
\label{se:six}

This paper presented a unified, computationally efficient, and fast distributed control framework that integrates DNMPC and CBFs to enable safe locomotion for multi-agent quadrupedal robots operating under external disturbances, rough terrain, and uncertain obstacle locations. The proposed DNMPC incorporates discrete-time CBF constraints for nonlinear SRB locomotion models to ensure both robot-to-robot and robot-to-obstacle safety. Additionally, a Lennard-Jones potential function is embedded in the NMPC cost formulation to facilitate consensus among agents. We validated the proposed approach through extensive hardware experiments on two A1 quadrupedal robots and numerical simulations involving a team of four A1 robots navigating uncertain environments with obstacle location uncertainty, rough terrain, and external disturbances. Our numerical results demonstrate a significant improvement in collision avoidance success rates compared to alternative DNMPC algorithms that rely solely on Euclidean distance constraints or apply CBFs at the WBC level.

In future work, we will explore integrating learning-based approaches with CBF-based DNMPC algorithms for agent foot placement to enhance the system’s adaptability to unforeseen challenges. Additionally, we will investigate the scalability of this framework to larger teams of quadrupedal robots navigating environments with a greater number of obstacles of increased complexity.




\vspace{-1em}
\bibliographystyle{IEEEtran}
\bibliography{references}

\begin{thebibliography}{10}
\providecommand{\url}[1]{#1}
\csname url@rmstyle\endcsname
\providecommand{\newblock}{\relax}
\providecommand{\bibinfo}[2]{#2}
\providecommand\BIBentrySTDinterwordspacing{\spaceskip=0pt\relax}
\providecommand\BIBentryALTinterwordstretchfactor{4}
\providecommand\BIBentryALTinterwordspacing{\spaceskip=\fontdimen2\font plus
\BIBentryALTinterwordstretchfactor\fontdimen3\font minus \fontdimen4\font\relax}
\providecommand\BIBforeignlanguage[2]{{%
\expandafter\ifx\csname l@#1\endcsname\relax
\typeout{** WARNING: IEEEtran.bst: No hyphenation pattern has been}%
\typeout{** loaded for the language `#1'. Using the pattern for}%
\typeout{** the default language instead.}%
\else
\language=\csname l@#1\endcsname
\fi
#2}}

\bibitem{Kim_Wensing_Convex_MPC_01}
J.~{Di Carlo}, P.~M. {Wensing}, B.~{Katz}, G.~{Bledt}, and S.~{Kim}, ``Dynamic locomotion in the {MIT Cheetah 3} through convex model-predictive control,'' in \emph{IEEE/RSJ International Conference on Intelligent Robots and Systems (IROS)}, Oct 2018, pp. 1--9.

\bibitem{Hutter_anymal_cbf_inWBC}
R.~Grandia, F.~Jenelten, S.~Yang, F.~Farshidian, and M.~Hutter, ``Perceptive locomotion through nonlinear model-predictive control,'' \emph{IEEE Transactions on Robotics}, vol.~39, no.~5, pp. 3402--3421, 2023.

\bibitem{MIT_humanoid_cbf_inWBC}
C.~Khazoom, S.~Hong, M.~Chignoli, E.~Stanger-Jones, and S.~Kim, ``Tailoring solution accuracy for fast whole-body model predictive control of legged robots,'' \emph{IEEE Robotics and Automation Letters}, vol.~9, no.~12, pp. 11\,074--11\,081, 2024.

\bibitem{CBF_MRS}
L.~Wang, A.~D. Ames, and M.~Egerstedt, ``Safety barrier certificates for collisions-free multirobot systems,'' \emph{IEEE Transactions on Robotics}, vol.~33, no.~3, pp. 661--674, 2017.

\bibitem{MRS_CBF_Cavorsi}
M.~Cavorsi, L.~Sabattini, and S.~Gil, ``Multirobot adversarial resilience using control barrier functions,'' \emph{IEEE Transactions on Robotics}, vol.~40, pp. 797--815, 2024.

\bibitem{Full_Koditschek_Template}
R.~Full and D.~Koditschek, ``{Templates and anchors: {N}euromechanical hypotheses of legged locomotion on land},'' \emph{Journal of Experimental Biology}, vol. 202, no.~23, pp. 3325--3332, 1999.

\bibitem{kajita19991LIP}
S.~Kajita and K.~Tani, ``Study of dynamic biped locomotion on rugged terrain-derivation and application of the linear inverted pendulum mode,'' in \emph{IEEE International Conference on Robotics and Automation}, 1991, pp. 1405--1406.

\bibitem{ALIP}
G.~Gibson, O.~Dosunmu-Ogunbi, Y.~Gong, and J.~Grizzle, ``Terrain-adaptive, {ALIP}-based bipedal locomotion controller via model predictive control and virtual constraints,'' in \emph{IEEE/RSJ International Conference on Intelligent Robots and Systems (IROS)}, 2022, pp. 6724--6731.

\bibitem{SLIP}
H.~Geyer, A.~Seyfarth, and R.~Blickhan, ``Compliant leg behavior explains basic dynamics of walking and running,'' \emph{Proceedings. Biological sciences / The Royal Society}, vol. 273, pp. 2861--7, 08 2006.

\bibitem{vLIP_Sreenath}
Z.~Li, J.~Zeng, S.~Chen, and K.~Sreenath, ``Autonomous navigation of underactuated bipedal robots in height-constrained environments,'' \emph{The International Journal of Robotics Research}, vol.~42, no.~8, pp. 565--585, 2023.

\bibitem{HLIP_Ames}
X.~Xiong and A.~Ames, ``3-{D} underactuated bipedal walking via {H-LIP} based gait synthesis and stepping stabilization,'' \emph{IEEE Transactions on Robotics}, vol.~38, no.~4, pp. 2405--2425, 2022.

\bibitem{orin2013centroidal}
D.~E. Orin, A.~Goswami, and S.-H. Lee, ``Centroidal dynamics of a humanoid robot,'' \emph{Autonomous robots}, vol.~35, no.~2, pp. 161--176, 2013.

\bibitem{Wensing_VBL_HJB}
M.~Chignoli and P.~M. Wensing, ``Variational-based optimal control of underactuated balancing for dynamic quadrupeds,'' \emph{IEEE Access}, vol.~8, pp. 49\,785--49\,797, 2020.

\bibitem{Abhishek_Hae-Won_TRO}
Y.~Ding, A.~Pandala, C.~Li, Y.-H. Shin, and H.-W. Park, ``Representation-free model predictive control for dynamic motions in quadrupeds,'' \emph{IEEE Transactions on Robotics}, vol.~37, no.~4, pp. 1154--1171, 2021.

\bibitem{pandala2022robust}
A.~Pandala, R.~T. Fawcett, U.~Rosolia, A.~D. Ames, and K.~Akbari~Hamed, ``Robust predictive control for quadrupedal locomotion: Learning to close the gap between reduced-and full-order models,'' \emph{IEEE Robotics and Automation Letters}, vol.~7, no.~3, pp. 6622--6629, 2022.

\bibitem{Leila_Hamed_RAL}
L.~Amanzadeh, T.~Chunawala, R.~T. Fawcett, A.~Leonessa, and K.~Akbari~Hamed, ``Predictive control with indirect adaptive laws for payload transportation by quadrupedal robots,'' \emph{IEEE Robotics and Automation Letters}, vol.~9, no.~11, pp. 10\,359--10\,366, 2024.

\bibitem{Basit_ASME}
B.~M. Imran, R.~T. Fawcett, J.~Kim, A.~Leonessa, and K.~Akbari~Hamed, ``{A Distributed Layered Planning and Control Algorithm for Teams of Quadrupedal Robots: An Obstacle-Aware Nonlinear MPC Approach},'' \emph{Journal of Dynamic Systems, Measurement, and Control}, vol. 147, no.~3, 2025.

\bibitem{NMPC_Park_02}
S.~Hong, J.-H. Kim, and H.-W. Park, ``Real-time constrained nonlinear model predictive control on {SO(3)} for dynamic legged locomotion,'' in \emph{IEEE/RSJ International Conference on Intelligent Robots and Systems (IROS)}, 2020, pp. 3982--3989.

\bibitem{NMPC_CBF_Ames_Hutter}
R.~Grandia, A.~J. Taylor, A.~D. Ames, and M.~Hutter, ``Multi-layered safety for legged robots via control barrier functions and model predictive control,'' in \emph{2021 IEEE International Conference on Robotics and Automation}, 2021, pp. 8352--8358.

\bibitem{NMPC_CBF_Sreenath}
Q.~Liao, Z.~Li, A.~Thirugnanam, J.~Zeng, and K.~Sreenath, ``Walking in narrow spaces: Safety-critical locomotion control for quadrupedal robots with duality-based optimization,'' in \emph{2023 IEEE/RSJ International Conference on Intelligent Robots and Systems (IROS)}, 2023, pp. 2723--2730.

\bibitem{Ames_CBF}
A.~D. Ames, X.~Xu, J.~W. Grizzle, and P.~Tabuada, ``Control barrier function based quadratic programs for safety critical systems,'' \emph{IEEE Transactions on Automatic Control}, vol.~62, no.~8, pp. 3861--3876, Aug 2017.

\bibitem{Nonsmooth_CBF}
A.~Thirugnanam, J.~Zeng, and K.~Sreenath, ``Nonsmooth control barrier functions for obstacle avoidance between convex regions,'' \emph{arXiv:2306.13259}, 2023.

\bibitem{JKim_CBF_CooperativeLoco}
J.~Kim, J.~Lee, and A.~D. Ames, ``Safety-critical coordination for cooperative legged locomotion via control barrier functions,'' in \emph{2023 IEEE/RSJ International Conference on Intelligent Robots and Systems (IROS)}, 2023, pp. 2368--2375.

\bibitem{Koushil_CBFMPC}
J.~Zeng, B.~Zhang, and K.~Sreenath, ``Safety-critical model predictive control with discrete-time control barrier function,'' in \emph{2021 American Control Conference (ACC)}, 2021, pp. 3882--3889.

\bibitem{CBFNMPC_Koushil}
J.~Zeng, Z.~Li, and K.~Sreenath, ``Enhancing feasibility and safety of nonlinear model predictive control with discrete-time control barrier functions,'' in \emph{2021 60th IEEE Conference on Decision and Control (CDC)}, 2021, pp. 6137--6144.

\bibitem{Jeeseop_TRO}
J.~Kim, R.~T. Fawcett, V.~R. Kamidi, A.~D. Ames, and K.~Akbari~Hamed, ``Layered control for cooperative locomotion of two quadrupedal robots: Centralized and distributed approaches,'' \emph{IEEE Transactions on Robotics}, vol.~39, no.~6, pp. 4728--4748, 2023.

\bibitem{Siljak_Decentralized_Book}
D.~Siljak, \emph{Decentralized Control of Complex Systems}.\hskip 1em plus 0.5em minus 0.4em\relax Dover Publications, December 2011.

\bibitem{2009_Scattolini_DMPC}
R.~Scattolini, ``Architectures for distributed and hierarchical model predictive control---{A} review,'' \emph{Journal of Process Control}, vol.~19, no.~5, pp. 723--731, 2009.

\bibitem{venkat2005stability}
A.~N. Venkat, J.~B. Rawlings, and S.~J. Wright, ``Stability and optimality of distributed model predictive control,'' in \emph{Proceedings of the IEEE Conference on Decision and Control}, 2005, pp. 6680--6685.

\bibitem{rawlings2008coordinating}
J.~B. Rawlings and B.~T. Stewart, ``Coordinating multiple optimization-based controllers: New opportunities and challenges,'' \emph{Journal of process control}, vol.~18, no.~9, pp. 839--845, 2008.

\bibitem{richards2007robust}
A.~Richards and J.~P. How, ``Robust distributed model predictive control,'' \emph{International Journal of control}, vol.~80, no.~9, pp. 1517--1531, 2007.

\bibitem{camponogara2002distributed}
E.~Camponogara, D.~Jia, B.~H. Krogh, and S.~Talukdar, ``Distributed model predictive control,'' \emph{IEEE control systems magazine}, vol.~22, no.~1, pp. 44--52, 2002.

\bibitem{Abdelaal2019}
M.~Abdelaal and S.~Sch{\"o}n, ``Distributed nonlinear model predictive control for connected vehicles trajectory tracking and collision avoidance with ellipse geometry,'' in \emph{Proceedings of the 32nd International Technical Meeting of the Satellite Division of The Institute of Navigation (ION GNSS+ 2019)}, 2019, pp. 2100--2111.

\bibitem{non-convex-PB-DMPC}
R.~Mao and H.~Dai, ``Distributed non-convex model predictive control for non-cooperative collision avoidance of networked differential drive mobile robots,'' \emph{IEEE Access}, vol.~10, pp. 52\,674--52\,685, 2022.

\bibitem{FARINA20121088}
M.~Farina and R.~Scattolini, ``Distributed predictive control: A non-cooperative algorithm with neighbor-to-neighbor communication for linear systems,'' \emph{Automatica}, vol.~48, no.~6, pp. 1088--1096, 2012.

\bibitem{Jessy_Book}
E.~Westervelt, J.~Grizzle, C.~Chevallereau, J.~Choi, and B.~Morris, \emph{Feedback Control of Dynamic Bipedal Robot Locomotion}.\hskip 1em plus 0.5em minus 0.4em\relax Taylor \& Francis/CRC, 2007.

\bibitem{Randy_ICRA_MultiAgent}
R.~T. Fawcett, L.~Amanzadeh, J.~Kim, A.~D. Ames, and K.~Akbari~Hamed, ``Distributed data-driven predictive control for multi-agent collaborative legged locomotion,'' in \emph{IEEE International Conference on Robotics and Automation (ICRA)}, 2023, pp. 9924--9930.

\bibitem{Multi_Agent_Quadrupeds_Sreenath}
C.~Yang, G.~N. Sue, Z.~Li, L.~Yang, H.~Shen, Y.~Chi, A.~Rai, J.~Zeng, and K.~Sreenath, ``Collaborative navigation and manipulation of a cable-towed load by multiple quadrupedal robots,'' \emph{IEEE Robotics and Automation Letters}, vol.~7, no.~4, pp. 10\,041--10\,048, 2022.

\bibitem{Borrelli_Opt_Coll_Avoi}
X.~Zhang, A.~Liniger, and F.~Borrelli, ``Optimization-based collision avoidance,'' \emph{IEEE Transactions on Control Systems Technology}, vol.~29, no.~3, pp. 972--983, 2021.

\bibitem{DT-HOCBF}
Y.~Xiong, D.-H. Zhai, M.~Tavakoli, and Y.~Xia, ``Discrete-time control barrier function: High-order case and adaptive case,'' \emph{IEEE Transactions on Cybernetics}, vol.~53, no.~5, pp. 3231--3239, 2023.

\bibitem{Khalil_Book}
H.~K. Khalil, \emph{Nonlinear Systems}.\hskip 1em plus 0.5em minus 0.4em\relax Pearson, 3rd edition, 2001.

\bibitem{LJ1924}
J.~E. Jones, ``On the determination of molecular fields. i. from the variation of the viscosity of a gas with temperature,'' \emph{Proceedings of The Royal Society A: Mathematical, Physical and Engineering Sciences}, vol. 106, pp. 441--462, 1924.

\bibitem{Randy_Paper_LCSS}
R.~T. Fawcett, A.~Pandala, A.~D. Ames, and K.~Akbari~Hamed, ``Robust stabilization of periodic gaits for quadrupedal locomotion via {QP}-based virtual constraint controllers,'' \emph{IEEE Control Systems Letters}, pp. 1736--1741, 2021.

\bibitem{MARTINS2022e01068}
O.~O. Martins, A.~A. Adekunle, O.~M. Olaniyan, and B.~O. Bolaji, ``An improved multi-objective a-star algorithm for path planning in a large workspace: Design, implementation, and evaluation,'' \emph{Scientific African}, vol.~15, p. e01068, 2022.

\bibitem{spong2005robot}
M.~Spong, S.~Hutchinson, and M.~Vidyasagar, \emph{Robot Modeling and Control}, ser. Wiley select coursepack.\hskip 1em plus 0.5em minus 0.4em\relax Wiley, 2005.

\bibitem{hybrid_A-star_APF}
J.~Wang, X.~Zhu, M.~Guo, S.~Yao, and Y.~Su, ``Improved hybrid algorithm of path planning for automated guided vehicle in storage system,'' in \emph{Proceedings of the International Conference on Control, Automation and Artificial Intelligence}.\hskip 1em plus 0.5em minus 0.4em\relax Atlantis Press, 2017, pp. 332--336.

\bibitem{obstacle_detector}
M.~Przybyła and A.~Milesi, ``obstacle\_detector,'' \url{https://github.com/tysik/obstacle_detector}, 2021, accessed: April 7, 2023.

\bibitem{RAISIM}
J.~{Hwangbo}, J.~{Lee}, and M.~{Hutter}, ``Per-contact iteration method for solving contact dynamics,'' \emph{IEEE Robotics and Automation Letters}, vol.~3, no.~2, pp. 895--902, April 2018.

\bibitem{CasADI}
J.~A.~E. Andersson, J.~Gillis, G.~Horn, J.~B. Rawlings, and M.~Diehl, ``{CasADi} -- {A} software framework for nonlinear optimization and optimal control,'' \emph{Mathematical Programming Computation}, vol.~11, no.~1, pp. 1--36, 2019.

\bibitem{YouTube_CBF_DNMPC}
``Safety-critical and distributed nonlinear predictive controllers for teams of quadrupedal robots. {[Online]. Available}: \url{https://youtu.be/N0z3zvkmvW4?si=TReWejO20OoOA7ND}.''

\bibitem{nguyen20163d}
Q.~{Nguyen}, A.~{Hereid}, J.~W. {Grizzle}, A.~D. {Ames}, and K.~{Sreenath}, ``{3D} dynamic walking on stepping stones with control barrier functions,'' in \emph{IEEE Conference on Decision and Control (CDC)}, Dec 2016, pp. 827--834.

\end{thebibliography}

\end{document}